\documentclass[runningheads]{llncs}

\usepackage{eccv}

\usepackage{eccvabbrv}

\usepackage{graphicx}
\usepackage{booktabs}

\usepackage[accsupp]{axessibility}  % Improves PDF readability for those with disabilities.

\usepackage{multirow}
\usepackage{makecell}
\usepackage[table]{xcolor}
\usepackage{colortbl}
\usepackage{amssymb}
\usepackage{subcaption}

\usepackage[pagebackref,breaklinks,colorlinks,citecolor=eccvblue]{hyperref}
\usepackage{orcidlink}

\begin{document}

% \title{Make 3D Tokens Count for Dynamic Spatial Reasoning}
% \title{Make 3D Count for Dynamic Spatial Reasoning}
% \title{Make 3D Tokens Count for Dynamic Reasoning}
% \title{Make Geometry Count for Dynamic Spatial Reasoning}
% \title{Forcing Dynamic Spatial Reasoning with 3D Tokens}
% \title{Make Geometry Matter for 4D Reasoning}
% \title{Geometric Spatial Reasoning: A Rethink}
% \title{Harnessing Geometry for Spatial Reasoning}

\title{Make Geometry Matter for Spatial Reasoning}

\titlerunning{GeoSR}

\author{Shihua Zhang \and
Qiuhong Shen \and
Shizun Wang \and
Tianbo Pan \and
Xinchao Wang\thanks{Corresponding author.}
}

\authorrunning{S.~Zhang et al.}

\institute{National University of Singapore, Singapore \\
\email{suhzhang001@gmail.com, qiuhong.shen@u.nus.edu, shizun.wang@u.nus.edu, e1583392@u.nus.edu, xinchao@nus.edu.sg}\\
}

\maketitle

\begin{abstract}
  Empowered by large-scale training, vision-language models (VLMs) achieve strong image and video understanding, yet their ability to perform spatial reasoning in both static scenes and dynamic videos remains limited. Recent advances try to handle this limitation by injecting geometry tokens from pretrained 3D foundation models into VLMs. Nevertheless, we observe that naive token fusion followed by standard fine-tuning in this line of work often leaves such geometric cues underutilized for spatial reasoning, as VLMs tend to rely heavily on 2D visual cues.
  In this paper, we propose \textbf{GeoSR}, a framework designed to make geometry matter by encouraging VLMs to actively reason with geometry tokens.
  \textbf{GeoSR} introduces two key components: (1) \textit{Geometry-Unleashing Masking}, which strategically masks portions of 2D vision tokens during training to weaken non-geometric shortcuts and force the model to consult geometry tokens for spatial reasoning; and (2) \textit{Geometry-Guided Fusion}, a gated routing mechanism that adaptively amplifies geometry token contributions in regions where geometric evidence is critical. Together, these designs unleash the potential of geometry tokens for spatial reasoning tasks.
  Extensive experiments on both static and dynamic spatial reasoning benchmarks demonstrate that \textbf{GeoSR} consistently outperforms prior methods and establishes new state-of-the-art performance by effectively leveraging geometric information.
  The project page is available at \url{https://suhzhang.github.io/GeoSR/}.
\end{abstract}

\section{Introduction}
\label{sec:intro}

Vision-language models (VLMs) that combine visual perception with language comprehension have advanced rapidly with large-scale training, showing strong capability in image and video understanding~\cite{yin2024survey,yu2025discrete,zhang2024vision,tang2026video,fengefficient,li2026sponge}. 
However, many real applications demand more than recognizing objects and events. They require \emph{spatial reasoning}, namely answering questions about where things are, how they relate in 3D space, and how these relations evolve over time.
This capability is essential in both \emph{static} settings, where scenes remain largely rigid but viewpoints and visibility vary, and \emph{dynamic} settings, where motion, occlusion, and temporal continuity play central roles. 
Recent benchmarks consistently reveal a significant gap: while prevalent VLMs excel at general visual semantics, they become brittle when faced with spatial questions involving viewpoint changes, motion continuity, and quantitative spatiotemporal judgments~\cite{tong2024cambrian,ma20253dsrbench,feng2025can,yang2025thinking,zhou2025vlm4d,li2025sti,zhu20254d,zhou2026learning}.

A natural direction to bridge this gap is introducing \emph{geometric cues} that provide 3D structural information for spatial reasoning.
Early 3D-aware MLLMs often rely on \emph{explicit} 3D inputs, such as depth signals~\cite{zhu2025llava,cheng2026sr3d}, point clouds~\cite{hong20233d,xu2024pointllm,zheng2025video,zhou2025llava}, or pre-built 3D maps~\cite{qi2025gpt4scene,xu2025chatbev}.
However, such pipelines typically require additional sensors or multi-stage 3D reconstruction processes, and the resulting geometry can be noisy, potentially degrading downstream performance and limiting scalability in the common scenarios where only monocular images or videos are available~\cite{zheng2025learning,wu2025spatial}.
Recent work has shifted toward a more scalable paradigm that extracts \emph{implicit} geometric features from monocular videos via pretrained visual geometry grounding models~\cite{wang2025vggt,wang2025continuous,wang2026pi}, and injects them into VLMs as geometry tokens to complement vanilla vision tokens~\cite{wu2025spatial,zheng2025learning,fan2026vlm,zhou2026learning}.
This geometry-token injection paradigm appears straightforward by fusing geometry tokens with visual tokens and then fine-tuning the model under spatial reasoning supervision. 
While these pretrained priors excel at visual geometry grounding and integrating them appears intuitively promising, a fundamental question remains unexplored. \textbf{Do geometry tokens actually help spatial reasoning, or are they merely dispensable side signals?}

\begin{figure}[t]
  \centering
  \begin{subfigure}[t]{0.49\linewidth}
    \centering
    \includegraphics[width=\linewidth, trim=0mm 8mm 0mm 0mm, clip]{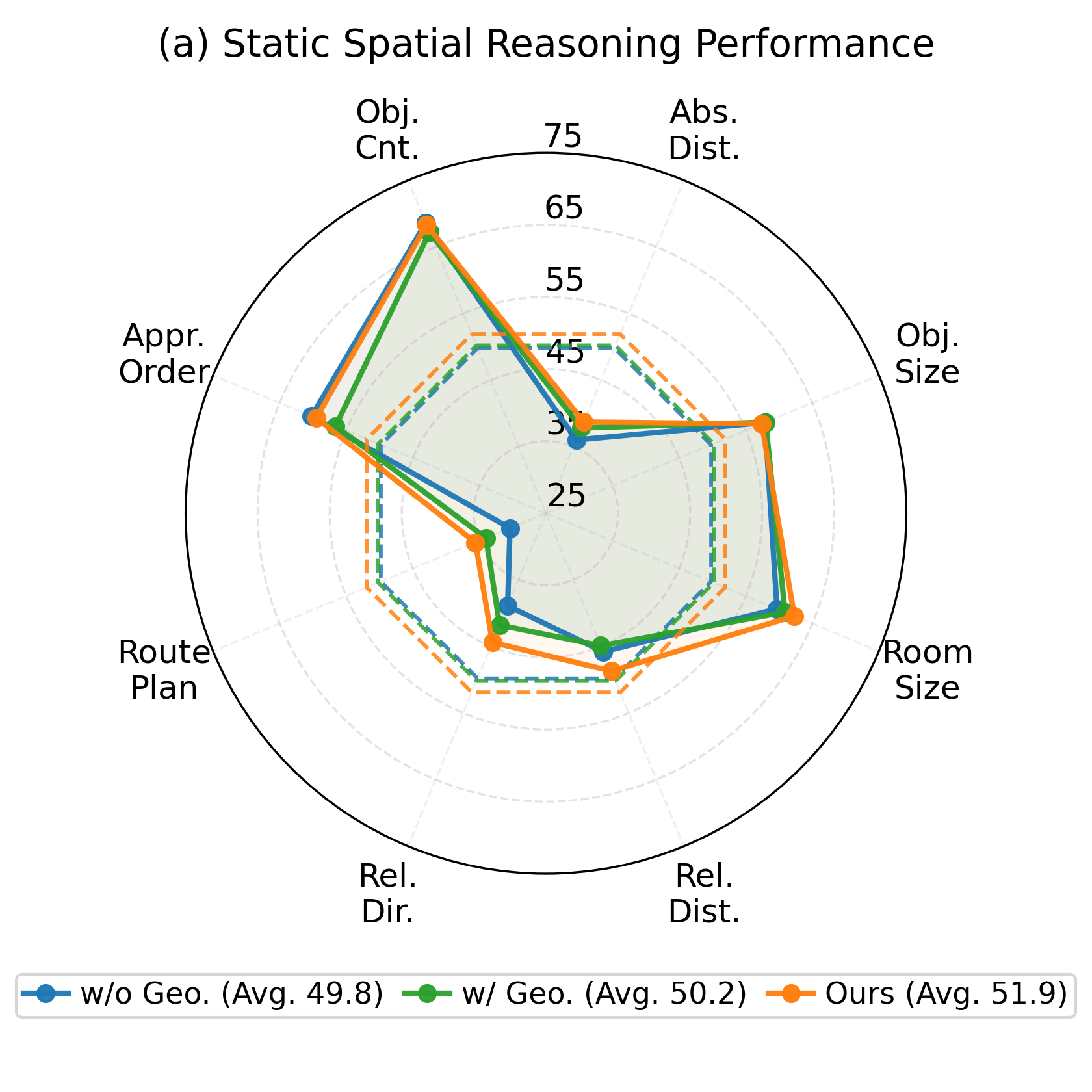}
    \label{fig:static_line}
  \end{subfigure}%\hfill
  \begin{subfigure}[t]{0.49\linewidth}
    \centering
    \includegraphics[width=\linewidth, trim=0mm 8mm 0mm 0mm, clip]{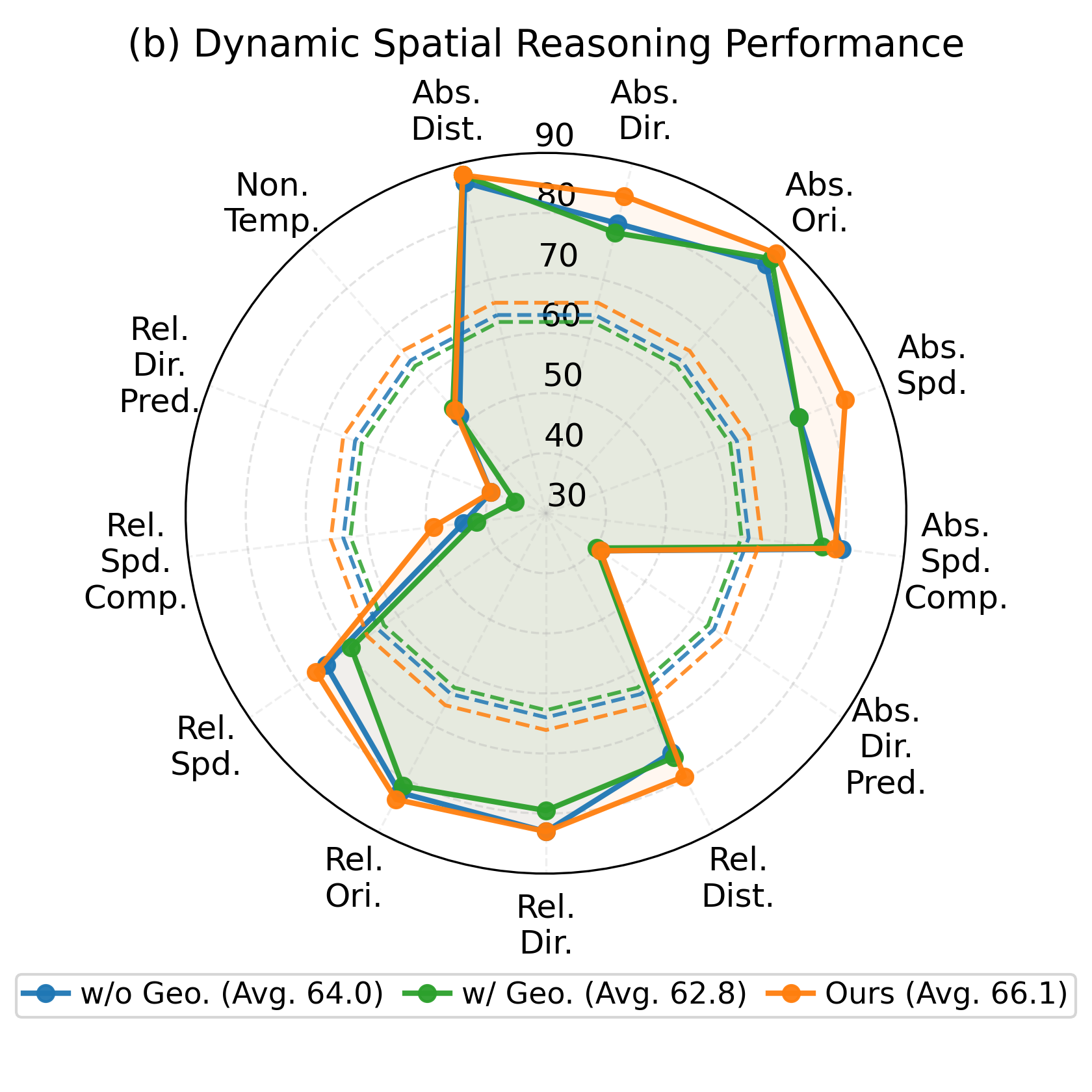}
    \label{fig:dynamic_line}
  \end{subfigure}
  % \vspace{-14pt}
  \caption{
    Geometry injection can be underutilized and even harmful for spatial reasoning. We compare three variants: w/o Geo. (removing the geometry branch), w/ Geo. (injecting geometry tokens via naive fusion with standard fine-tuning), and Ours. The plots reveal a counterintuitive yet reproducible pattern: in static scenes, w/ Geo. brings only marginal gains over w/o Geo., whereas in dynamic videos, w/ Geo. can even underperform the w/o Geo. baseline. This suggests that VLMs often fall back to appearance-driven shortcuts in 2D visual tokens and treat geometry as a dispensable side signal. These observations motivate \textbf{GeoSR} (Ours), which compels models to use geometry as actionable evidence and yields consistent improvements across both static and dynamic spatial reasoning.
  }
  \label{fig:intro}
\end{figure}

Surprisingly, we observe a counterintuitive but reproducible phenomenon that under the common practice of naive token fusion followed by standard fine-tuning, geometry tokens are often underutilized for spatial reasoning. 
As illustrated in Figure~\ref{fig:intro}, for static scenes, performance gains can be limited, and for dynamic scenes, injecting geometry tokens may even yield negative effects. This suggests that the models frequently default to appearance-driven shortcuts in 2D vision tokens, treating geometry as a dispensable auxiliary signal, rather than as actionable evidence for spatial reasoning.
Moreover, indiscriminately mixing geometry with appearance may dilute useful geometric cues, thus the model still avoids consulting geometry when appearance shortcuts are barely sufficient.
Clearly, it is insufficient to merely inject geometry tokens naively. 
\textbf{We must compel the VLM to effectively rely on geometric information when spatial reasoning demands it.}

Motivated by this principle, we propose \textbf{GeoSR}, a framework that makes geometry matter for spatial reasoning in both static and dynamic settings.
\textbf{GeoSR} addresses the problem from two complementary angles.
First, to make geometry usage \emph{effective}, we introduce \textbf{Geometry-Unleashing Masking}. During training, it strategically masks parts of the 2D vision tokens to weaken non-geometric shortcuts, so the model is encouraged to consult geometry tokens to answer spatial questions.
Second, to make geometry usage \emph{reasonable}, we propose \textbf{Geometry-Guided Fusion}, a gated routing mechanism that adaptively increases the contribution of geometry tokens at regions where geometric evidence is needed, instead of treating geometry as uniformly helpful across all tokens and frames.
Together, these designs make geometry tokens actionable and controllable, rather than being easily ignored or indiscriminately fused.

In summary, our main contributions are as follows.

\begin{itemize}
  \item We report a reproducible finding in spatial reasoning that implicit geometry injection can be ineffective under naive fusion and standard fine-tuning. Geometry tokens may be largely ignored, with limited or even negative gains on spatial reasoning in practice.
  \item We propose \textbf{GeoSR}, a simple framework that encourages effective and reasonable use of geometry tokens. Geometry-Unleashing Masking reduces reliance on appearance shortcuts, and Geometry-Guided Fusion routes geometry information adaptively when geometric evidence is required.
  \item We validate \textbf{GeoSR} on both static and dynamic spatial reasoning benchmarks, showing consistent improvements over prior methods.
\end{itemize}

% Extensive experiments on 4D reasoning benchmarks and general video understanding benchmarks show that \textbf{GeoSR} substantially improves 4D reasoning while maintaining competitive general video performance. Moreover, applying our strategy to existing static spatial reasoning methods also yields consistent gains.

\section{Related Work}
\label{sec:related}

\subsection{General Video Understanding}

Recent advances have extended vision-language models from static images to videos by representing a clip as multiple sampled frames and modeling temporal context for tasks such as video captioning~\cite{maaz2024video,zhang2023video} and VideoQA~\cite{zhang2025llava,yuan2025videorefer}.
Proprietary video-capable VLMs provide strong upper-bound baselines for general video understanding~\cite{hurst2024gpt,singh2025openai,comanici2025gemini}.
In parallel, several open-source models are explicitly tailored for video understanding, typically enhancing temporal modeling and long-context handling for multi-frame inputs~\cite{zhang2025llava,yuan2025videorefer,chen2025scaling}.
Meanwhile, general-purpose VLMs have also been rapidly extended to video inputs via multi-frame prompting and instruction tuning, offering broad coverage across diverse video tasks~\cite{bai2025qwen25vl,bai2025qwen3,wang2025internvl3}.
Despite strong performance on general video understanding, these models are often brittle on spatial reasoning~\cite{yang2025thinking,ma20253dsrbench,zhou2026learning}.
This is because common supervision mainly rewards semantic alignment, making models unstable under viewpoint changes, occlusion, and motion, thus requiring geometry-grounded evidence for reliable spatial reasoning~\cite{yang2025thinking,ma20253dsrbench,zhou2026learning}.
% In contrast to improving general video semantics, our work focuses on making injected geometry evidence actionable for spatial reasoning via shortcut suppression and controlled fusion.

\subsection{Spatial Reasoning with VLMs}

Spatial reasoning aims to answer questions about spatial layout and measurable relations, such as distance, direction, and motion changes over time.
Recent benchmarks show that such queries remain challenging for modern VLMs, especially when viewpoints vary or relations evolve dynamically~\cite{yang2025thinking,ma20253dsrbench,zhou2026learning,zhou2025vlm4d,li2025sti,zhu20254d}.

\noindent \textbf{Static scenes.}
% \paragraph{Static Scenes.}
One of the common settings are reasoning in static scenes, where the environment is largely rigid while camera viewpoint and visibility vary across frames.
VSI-Bench~\cite{yang2025thinking} is a representative benchmark for this regime and highlights the need for viewpoint-robust spatial understanding beyond frame-specific appearance cues.
To strengthen spatial capability, one line of work scales spatial supervision through large spatial question-answer (QA) pairs and instruction tuning~\cite{zhang2025from,ray2025sat}.
Another line of work injects geometric priors from pretrained 3D foundation models~\cite{wang2025vggt,wang2026pi,wang2025continuous} and fuses them with 2D vision tokens~\cite{wu2025spatial,zheng2025learning,fan2026vlm}.
VG-LLM~\cite{zheng2025learning} extracts implicit geometry tokens from~\cite{wang2025vggt} and performs token-level fusion to improve spatial reasoning under viewpoint change.
Spatial-MLLM~\cite{wu2025spatial} further argues that conventional semantic encoders under-emphasize structure, and proposes a spatial branch with space-aware frame sampling to better preserve geometry-relevant evidence under limited video context.
VLM-3R~\cite{fan2026vlm} augments VLMs with instruction-aligned 3D reconstruction priors and integrates reconstruction-aware representations to improve static spatial reasoning.
While these methods demonstrate the promise of geometry-aware enhancements, they often rely on uniform or naive fusion, which can leave geometry cues underutilized.
Our work targets this gap by making injected geometry evidence more actionable during reasoning. %, via the proposed Geometry-Unleashing Masking and Geometry-Guided Fusion.

\noindent \textbf{Dynamic scenes.}
% \paragraph{Dynamic Scenes.}
Dynamic spatial reasoning, also referred to as 4D reasoning, considers scenarios where spatial relations change over time due to camera motion and object motion. Correct answers require spatiotemporal consistency rather than single-frame cues~\cite{zhou2026learning}.
As an early attempt toward 4D scene understanding, LLaVA-4D~\cite{zhou2025llava} embeds spatiotemporal prompts into MLLMs to reason about evolving object states in 4D settings.
Most closely related, GSM~\cite{zhou2026learning} retrieves question-relevant dynamic geometric evidence from pretrained dynamic geometry priors~\cite{wang2026pi}, and appends them to vision tokens as the input of VLM.
All these methods try to inject geometry cues to improve 4D reasoning. However, the former requires multiple videos as input and relies on explicit SfM-style reconstruction~\cite{schonberger2016structure,zhang2025deep}, which limits scalability in monocular in-the-wild videos. The latter actually shows limited gains from geometry injection, which will be discussed in our paper, highlighting that indiscriminate geometry fusion can be ineffective or even harmful without proper control.
In contrast, our work focuses on making geometry tokens from monocular videos actionable for spatial reasoning.

\section{Method}
\label{sec:method}

In this section, we first review the commonly used geometry injection framework for spatial reasoning, which augments a VLM with a pretrained geometry branch and a fusion module.
We then present two key strategies in our \textbf{GeoSR}, \textbf{Geometry-Unleashing Masking} and \textbf{Geometry-Guided Fusion}, to make geometry tokens actionable rather than ignorable.
Finally, we provide implementation details for both static and dynamic spatial reasoning.

\subsection{Preliminary: Geometry-Aware Framework}
\label{subsec:framework}

\begin{figure}[t]
    \centering
    \includegraphics[width=0.99\linewidth]{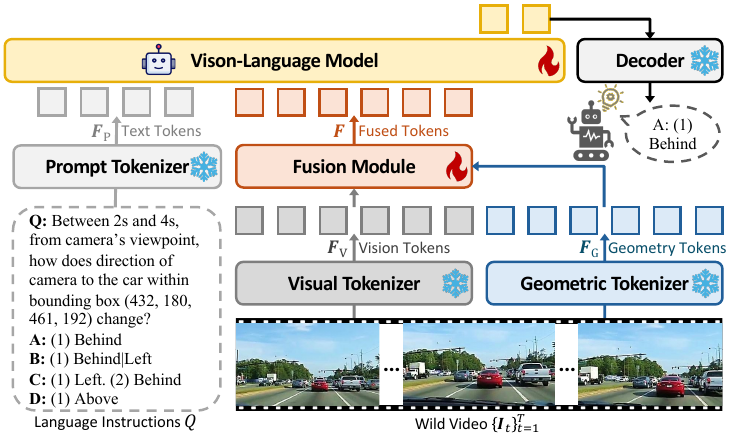}
    \caption{
      Overview of the Geometry-Aware Framework for spatial reasoning.
      It augments a VLM with an additional geometry branch.
      A pretrained geometric tokenizer extracts geometry tokens $\boldsymbol{F}_{\mathrm{G}}$ from the input video, which are fused with standard vision tokens $\boldsymbol{F}_{\mathrm{V}}$ by a fusion module to form $\boldsymbol{F}$.
      The VLM answers the query based on the fused evidence together with the text tokens $\boldsymbol{F}_{\mathrm{P}}$.
      Snow and flame icons denote frozen and trainable components, respectively.
      This paradigm serves as our baseline.
      Our key observation is that naive fusion under standard fine-tuning can leave $\boldsymbol{F}_{\mathrm{G}}$ underutilized, which motivates \textbf{GeoSR} to make geometry really matter for spatial reasoning.
    }
    \label{fig:framework}
\end{figure}

Traditional VLMs have to infer the 3D structure and dynamics of the scene from only 2D vision tokens for text-prompted spatial reasoning, which can be challenging and may lead to suboptimal performance. Accordingly, recent methods attempt to improve the geometric awareness of VLMs by introducing geometry tokens, forming a widely-used geometry-aware framework.
The overall architecture of this framework is illustrated in Figure~\ref{fig:framework}.
Given a sequence of images or video frames $\{\boldsymbol{I}_t\vert \boldsymbol{I}_t\in\mathbb{R}^{H_I\times W_I\times 3}\}_{t=1}^T$ and a text prompt $Q$, the goal is to answer the question based on the visual and textual information.
For the tokenizer, except for the vision branch that encodes $\{\boldsymbol{I}_t\}_{t=1}^T$ to 2D tokens $\boldsymbol{F}_{\text{V}}\in\mathbb{R}^{(H_{\text{V}}W_{\text{V}}T)\times C_{\text{V}}}$ and the prompt branch that encodes $Q$ to text tokens $\boldsymbol{F}_{\text{P}}\in\mathbb{R}^{L_{\text{P}}\times C_{\text{P}}}$, it introduces an additional geometry branch that extracts implicit geometry cues from $\{\boldsymbol{I}_t\}_{t=1}^T$ typically using a pretrained 3D model~\cite{wang2025vggt,wang2025continuous,wang2026pi}. The geometry tokens $\boldsymbol{F}_{\text{G}}\in\mathbb{R}^{(H_{\text{G}}W_{\text{G}}T)\times C_{\text{G}}}$ are then merged with the 2D vision tokens $\boldsymbol{F}_{\text{V}}$ in a fusion module $\mathcal{F}(\cdot)$, and the fused tokens $\boldsymbol{F}\in\mathbb{R}^{L\times C}$ are fed into the VLM backbone together with $\boldsymbol{F}_{\text{P}}$, generating the answer to the question. Generally, $C_{\text{V}}=C_{\text{P}}=C$ are the same, while $C_{\text{G}}$ can be different.
And the fusion module $\mathcal{F}(\cdot)$ often varies across different methods. We specify this in the following.

\noindent (1) \emph{Additive Fusion} is commonly used in static spatial reasoning~\cite{zheng2025learning}:
\begin{equation}
  \boldsymbol{F}=\boldsymbol{F}_{\text{V}}+\mathrm{MLP}(\mathrm{Reshape}(\boldsymbol{F}_{\text{G}})),
  \label{eq:fusion_static}
\end{equation}
where $\mathrm{Reshape}(\cdot)$ and $\mathrm{MLP}(\cdot)$ are used to align the geometry tokens with the vision tokens in terms of resolution and dimensionality~\cite{zheng2025learning}.

\noindent (2) \emph{QFormer fusion}, a more sophisticated module, is often needed to capture the complex spatiotemporal dynamics~\cite{zhou2026learning}. It first summarizes the text prompt's intent using a small set of learnable bottleneck tokens $\boldsymbol{B}\in\mathbb{R}^{L_{\text{B}}\times C}$ with cross-attention~\cite{vaswani2017attention}:
\begin{equation}
\boldsymbol{F}_{\text{B}}=\mathrm{CrossAttn}_1(\boldsymbol{B},\boldsymbol{F}_{\text{P}}),
\label{eq:pcm_bottleneck}
\end{equation}
where $\boldsymbol{B}$ acts as the query set and $\boldsymbol{F}_{\text{P}}$ serves as the key and value sets. $L_{\text{B}}$ is the length of bottleneck tokens, which is typically small to avoid high computational cost.
Then \emph{QFormer} queries these tokens among geometry tokens $\boldsymbol{F}_{\text{G}}$:
\begin{equation}
\boldsymbol{Z}_{\text{G}}
=\mathrm{CrossAttn}_2(\boldsymbol{F}_{\text{B}},\hat{\boldsymbol{F}}_{\text{G}})=
\mathrm{CrossAttn}_2(\boldsymbol{F}_{\text{B}},\mathrm{MLP}(\boldsymbol{F}_{\text{G}})),
\label{eq:pcm_crossattn_geo_old}
\end{equation}
$\boldsymbol{F}_{\text{B}}$ is the query set, and $\boldsymbol{F}_{\text{G}}\in \mathbb{R}^{(H_{\text{G}}W_{\text{G}}T)\times C_{\text{G}}}$ is projected to $\hat{\boldsymbol{F}}_{\text{G}}\in \mathbb{R}^{(H_{\text{G}}W_{\text{G}}T)\times C}$ with $\mathrm{MLP}(\cdot)$ to match the dimensionality, serving as the key and value sets. The output $\boldsymbol{Z}_{\text{G}}$ is full of geometry cues, which is appended to the vision tokens:
\begin{equation}
\boldsymbol{F}=[\boldsymbol{F}_{\text{V}}, \mathrm{MLP}(\boldsymbol{Z}_{\text{G}})],
\label{eq:fusion_dynamic}
\end{equation}
where $[\cdot, \cdot]$ denotes concatenation along the token dimension. 

Note that for these spatiotemporal inputs, tokens are typically equipped with spatiotemporal positional encodings, such as Rotary Position Embedding (RoPE)~\cite{su2024roformer} and its spatiotemporal extensions~\cite{wang2024qwen2,gao2025tc}, which we omit for clarity.
In practice, all prompt, visual, and geometric tokenizers are typically kept frozen, while the fusion module $\mathcal{F}(\cdot)$ is trained and the VLM backbone is fine-tuned on spatial reasoning datasets~\cite{wu2025spatial,zheng2025learning,fan2026vlm,zhou2026learning}.
% Albeit straightforward, we find that this common recipe does not unleash the potential of geometry tokens for spatial reasoning. The VLM can still rely on 2D visual cues and ignore the geometry stream, which leads to suboptimal performance and even the counterintuitive observation that removing the geometry branch further improves performance, as shown in Figure~\ref{fig:intro}.
Albeit straightforward, we find that this common recipe does not make geometry tokens actively useful for spatial reasoning. The VLM can still rely on 2D visual cues and largely ignore the geometry stream, leading to suboptimal performance. And in some cases, performance even improves after removing the geometry branch, as shown in Fig.~\ref{fig:intro}.
To address this counterintuitive issue, we propose \textbf{GeoSR} with two strategies to encourage the VLM to reason with geometry cues, which will be introduced in the following sections.

\subsection{Geometry-Unleashing Masking}
\label{subsec:masking}

\begin{figure}[t]
    \centering
    \includegraphics[width=0.99\linewidth]{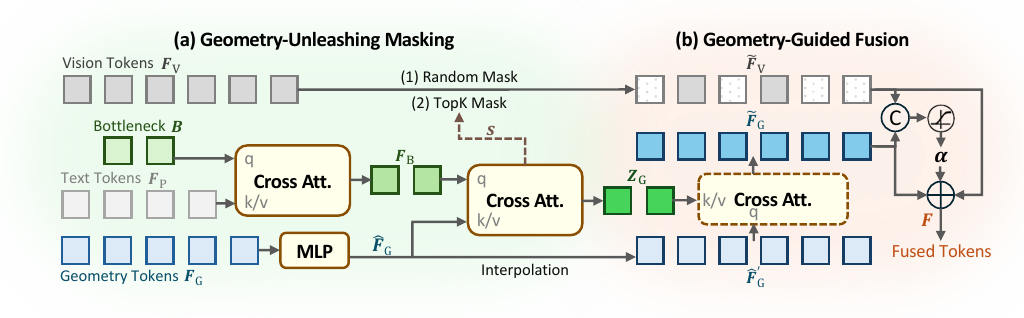}
    \caption{
      Overview of the proposed strategies in \textbf{GeoSR}.
      \textbf{(a) Geometry-Unleashing Masking} suppresses appearance shortcuts during training by masking a subset of vision tokens $\boldsymbol{F}_{\mathrm{V}}$.
      For static settings, the mask is sampled randomly.
      For dynamic settings, bottleneck tokens $\boldsymbol{B}$ first attend to text tokens $\boldsymbol{F}_{\mathrm{P}}$ to obtain $\boldsymbol{F}_{\mathrm{B}}$, which then attend to geometry tokens $\boldsymbol{F}_{\mathrm{G}}$ to produce question-relevant geometry evidence $\boldsymbol{Z}_{\mathrm{G}}$ and a relevance score used for TopK masking.
      \textbf{(b) Geometry-Guided Fusion} redistributes the compact evidence $\boldsymbol{Z}_{\mathrm{G}}$ back to token-level geometry features (if applicable) and applies a learned gate $\boldsymbol{\alpha}$ to control the contributions of masked vision features $\tilde{\boldsymbol{F}}_{\mathrm{V}}$ and geometry features $\tilde{\boldsymbol{F}}_{\mathrm{G}}$, producing fused tokens $\boldsymbol{F}$ for the VLM backbone.
    }
    \label{fig:module}
\end{figure}

As discussed in Section~\ref{sec:intro}, the core challenge is that the geometry-aware VLM may still answer spatial reasoning questions from 2D appearance shortcuts, even when geometry tokens are provided.
We therefore introduce Geometry-Unleashing Masking to suppress the shortcuts during training and force the model to effectively use geometric evidence.
The detailed architecture of this is illustrated in Figure~\ref{fig:module}(a), where the masking process can be formulated as:
\begin{equation}
\tilde{\boldsymbol{F}}_{\text{V}}
=
\boldsymbol{m}\odot \boldsymbol{F}_{\text{V}},
\label{eq:gum_apply}
\end{equation}
\begin{equation}
m_j=
\begin{cases}
0, & \text{if } \delta=1 \ \wedge \ j\in\mathcal{M},\\
1, & \text{otherwise},
\end{cases}
\quad \delta\sim\mathrm{Bernoulli}(\beta),
\label{eq:gum_enable}
\end{equation}
where $\boldsymbol{m}=\{m_j\}\in\{0,1\}^{(H_{\text{V}}W_{\text{V}}T)}$ is a binary mask over vision tokens, $\mathcal{M}$ is the mask set, $\beta$ controls the probability of enabling masking during training, and $\odot$ denotes element-wise multiplication.
So, now the question is how to define $\boldsymbol{m}$ in Eq.~\eqref{eq:gum_apply}.
A simple idea is to randomly mask vision tokens, such as MAE~\cite{he2022masked}, which has been proven to be effective in many vision tasks~\cite{tong2022videomae,kirillov2023segment,weinzaepfel2023croco}.
Inspired by this, \textbf{GeoSR} uniformly samples $\mathcal{M}$ of size $K$ from visual-token positions whose resolution is equal to $H_{\text{V}}W_{\text{V}}T$:
\begin{equation}
\mathcal{M}=\mathrm{Rand}(H_{\text{V}}W_{\text{V}}T,K),\quad
K=\lceil \gamma (H_{\text{V}}W_{\text{V}}T) \rceil,
\label{eq:gum_random}
\end{equation}
where $\gamma$ is the masking ratio.
Random masking is a simple yet effective choice for \textbf{static spatial reasoning} baselines that use \emph{Additive Fusion} as in Eq.~\eqref{eq:fusion_static}~\cite{wu2025spatial,zheng2025learning,fan2026vlm}.
Meanwhile, for \textbf{dynamic spatial reasoning} baselines that use \emph{QFormer Fusion} as in Eqs.~\eqref{eq:pcm_bottleneck} to~\eqref{eq:fusion_dynamic}~\cite{zhou2026learning}, \textbf{GeoSR} instead selects mask regions under this query-style mechanism.
With Eqs.~\eqref{eq:pcm_bottleneck} and~\eqref{eq:pcm_crossattn_geo_old} that effectively condense the question-related geometric information into $\boldsymbol{Z}_{\text{G}}$, we can identify the most relevant 3D tokens for answering the question, ranking all tokens according to the relevance score $\boldsymbol{s}\in {[0,1]}^{(H_{\text{G}}W_{\text{G}}T)}$. Then with the score, we select the positions of the most critical geometry tokens and mask the corresponding 2D vision tokens during training, which encourages the model to seek these important geometric cues from the geometry stream instead of relying on 2D appearance shortcuts.
We use the attention weights in Eq.~\eqref{eq:pcm_crossattn_geo_old} to derive the relevance score $\boldsymbol{s}$ for masking. This process can be formulated as:
\begin{equation}
\boldsymbol{Z}_{\text{G}},\boldsymbol{A}
=\mathrm{CrossAttn}_2(\boldsymbol{F}_{\text{B}},\hat{\boldsymbol{F}}_{\text{G}})=
\mathrm{CrossAttn}_2(\boldsymbol{F}_{\text{B}},\mathrm{MLP}(\boldsymbol{F}_{\text{G}})),
\label{eq:pcm_crossattn_geo}
\end{equation}
\begin{equation}
\boldsymbol{s}=\{s_j\}=\left\{\frac{u_j-\min(\boldsymbol{u})}{\max(\boldsymbol{u})-\min(\boldsymbol{u})+\epsilon}\right\},\quad
u_j=\frac{1}{hL_{\text{B}}}\sum_{k=1}^{h}\sum_{i=1}^{L_{\text{B}}}{A_{k,i,j}}.
\label{eq:pcm_score}
\end{equation}
Here, $\boldsymbol{A}=\{A_{k,i,j}\}\in{[0,1]}^{h\times L_{\text{B}}\times (H_{\text{G}}W_{\text{G}}T)}$ is the attention probability tensor, where $h$ is the number of attention heads. $\boldsymbol{u}_i$ is the average attention probability of each geometry token across all heads and bottleneck tokens, regarding as the relevance measurement, and we further normalize $\boldsymbol{u}$ to get the final relevance score $\boldsymbol{s}$. $\epsilon$ is a small value to ensure numerical stability.
Ideally, after obtaining $\boldsymbol{s}$, we can directly identify the mask set on high-relevant locations with TopK selection:
\begin{equation}
\mathcal{M}=\mathrm{TopK}(\boldsymbol{s},K),\quad K=\lceil \gamma (H_{\text{G}}W_{\text{G}}T)\rceil,
\label{eq:pcm_topk}
\end{equation}
which replaces the random mask in Eq.~\eqref{eq:gum_random}, and then perform masking process in Eqs.~\eqref{eq:gum_apply} and~\eqref{eq:gum_enable} for dynamic spatial reasoning.
However, there is one more issue to be considered. Note that $\boldsymbol{m}$ in Eq~\eqref{eq:gum_apply} should satisfy the resolution of ${H_{\text{V}}\times W_{\text{V}}\times T}$, which is the same as $\boldsymbol{F}_{\text{V}}$, while $\mathcal{M}$ is obtained from $\boldsymbol{F}_{\text{G}}$ with the resolution of $H_{\text{G}}\times W_{\text{G}}\times T$. The resolution may mismatch, silimar to Eq.~\eqref{eq:fusion_static}.
Therefore, the same 2D spatial interpolation may be needed here if the geometry tokens and vision tokens have different spatial resolutions:
   \begin{equation}
\hat{\boldsymbol{F}}_{\text{G}}'=\mathrm{Interp}(\hat{\boldsymbol{F}}_{\text{G}},(H_{\text{V}},W_{\text{V}}))=\mathrm{MLP}(\mathrm{Reshape}(\hat{\boldsymbol{F}}_{\text{G}})).
\label{eq:pcm_geo_align}
\end{equation}
In this case, $\hat{\boldsymbol{F}}_{\text{G}}'$ will replace $\hat{\boldsymbol{F}}_{\text{G}}$ in Eq.~\eqref{eq:pcm_crossattn_geo} while others remain unchanged.

From Eqs.~\eqref{eq:gum_apply} to~\eqref{eq:pcm_geo_align}, the Geometry-Unleashing Masking attempts to suppress some of the vision tokens during training, thus forcing the model to unleash geometry from the 3D branch, making effective geometric information utilization.
Then, the model has to fuse these geometric cues into the VLM backbone reasonably, fully leveraging their capabilities for both static and dynamic spatial reasoning.

\subsection{Geometry-Guided Fusion}
\label{subsec:fusion}

% Geometry-Unleashing Masking suppresses appearance-based shortcuts during training. Meanwhile, we should also fuse the geometric cues with the vision tokens and feed them into the VLM backbone.
Building on the above discussions, we next fuse the geometric cues with the visual tokens and feed the fused representations into the VLM backbone.
However, uniformly mixing in Eq.~\eqref{eq:fusion_static}~\cite{zheng2025learning,wu2025spatial} or simply concatenating in Eq.~\eqref{eq:fusion_dynamic}~\cite{zhou2026learning} still leaves the model an easy option to down-weight geometry everywhere and fall back to the remaining visual evidence. \textbf{GeoSR} therefore introduces Geometry-Guided Fusion, as shown in Figure~\ref{fig:module}(b), which adaptively controls how much geometric evidence is injected into each fused token in a fine-grained manner, ensuring that geometry dominates when it is informative and necessary.

% Generally, this designed module fuses the geometry features $\tilde{\boldsymbol{F}}_{\text{G}}=\hat{\boldsymbol{F}}_{\text{G}}'$ with the vision tokens $\tilde{\boldsymbol{F}}_{\text{V}}$ that will be masked during training (Eq.~\eqref{eq:gum_apply}) adaptively under the guidance of a learned gate mechanism, which allows the model to increase reliance on geometry when visual cues are suppressed and geometry is informative, rather than treating geometry as uniformly helpful. The gated fusion weights are computed as follows:
Generally, this designed module fuses the geometry features $\tilde{\boldsymbol{F}}_{\text{G}}$ (as defined in following Eqs.~\eqref{eq:ggf_crossattn_retrieve_static} and~\eqref{eq:ggf_crossattn_retrieve}) with the vision tokens $\tilde{\boldsymbol{F}}_{\text{V}}$ that will be masked during training (as in Eq.~\eqref{eq:gum_apply}) adaptively under the guidance of a learned gate mechanism, which allows the model to increase reliance on geometry when visual cues are suppressed and geometry is informative, rather than treating geometry as uniformly helpful. The gated fusion weights are computed as follows:
\begin{equation}
\boldsymbol{\alpha}=\sigma\!\left(\mathbf{W}_{g}[\boldsymbol{V}\Vert\boldsymbol{G}]+\boldsymbol{b}_{g}\right),\quad
\boldsymbol{V}=\mathrm{LN}_{v}(\tilde{\boldsymbol{F}}_{\text{V}}),\quad
\boldsymbol{G}=\mathrm{LN}_{g}(\tilde{\boldsymbol{F}}_{\text{G}}),
\label{eq:ggf_norm}
\end{equation}
where $[\cdot\Vert \cdot]$ denotes concatenation along the channel dimension, $\mathbf{W}_{g}$ and $\boldsymbol{b}_{g}$ are learnable parameters, and $\sigma(\cdot)$ is the sigmoid function. $\mathrm{LN}(\cdot)$ here denotes layer normalization~\cite{ba2016layer}. This operation outputs $\boldsymbol{\alpha}\in(0,1)^{(H_{\text{V}}W_{\text{V}}T)\times C}$, which is a token-and-channel-wise gate, enabling fine-grained routing of geometry evidence.
For \textbf{static spatial reasoning}, the geometry feature is defined as:
\begin{equation}
\tilde{\boldsymbol{F}}_{\text{G}}=\mathrm{MLP}(\mathrm{Reshape}(\boldsymbol{F}_{\text{G}})),
\label{eq:ggf_crossattn_retrieve_static}
\end{equation}
% as the same as Eq.~\eqref{eq:fusion_static}.
While for \textbf{dynamic spatial reasoning} that conducts Eq.~\eqref{eq:pcm_crossattn_geo}, the compact geometry evidence $\boldsymbol{Z}_{\text{G}}$ is first redistributed back to the full spatiotemporal geometry tokens $\hat{\boldsymbol{F}}_{\text{G}}'$ aligned with the vision tokens (as in Eq.~\eqref{eq:pcm_geo_align}). Then, Similar to Eqs.~\eqref{eq:pcm_bottleneck} and~\eqref{eq:pcm_crossattn_geo}, we use cross-attention to retrieve location-aware fine-grained geometry features:
% Additionally, for dynamic spatial reasoning that conducts Eq.~\eqref{eq:pcm_crossattn_geo}, the compact geometry evidence $\boldsymbol{Z}_{\text{G}}$ is first redistributed back to the full spatiotemporal geometry tokens $\hat{\boldsymbol{F}}_{\text{G}}'$ aligned with the vision tokens (as mentioned in Eq.~\eqref{eq:pcm_geo_align}). Then, Similar to Eqs.~\eqref{eq:pcm_bottleneck} and~\eqref{eq:pcm_crossattn_geo}, we use cross-attention to retrieve location-aware fine-grained geometry features:
\begin{equation}
\tilde{\boldsymbol{F}}_{\text{G}}
=
\mathrm{CrossAttn}_3\!\left(\hat{\boldsymbol{F}}_{\text{G}}',\boldsymbol{Z}_{\text{G}}\right),
\label{eq:ggf_crossattn_retrieve}
\end{equation}
where $\hat{\boldsymbol{F}}_{\text{G}}'$ is the query set, and $\boldsymbol{Z}_{\text{G}}\in\mathbb{R}^{L_{\text{B}}\times C}$ serves as the key and value sets. The output $\tilde{\boldsymbol{F}}_{\text{G}}\in\mathbb{R}^{(H_{\text{V}}W_{\text{V}}T)\times C}$ is a fine-grained geometry feature map then fed to Eq.~\eqref{eq:ggf_norm}.
This retrieval step emphasizes the dynamics condensed in $\boldsymbol{Z}_{\text{G}}$.
% This retrieval step ensures that each spatial-temporal position can access the geometry evidence summarized in $\boldsymbol{Z}_{\text{G}}$ where the evidence is highly relevant to the text.

Finally, after Eq.~\eqref{eq:ggf_norm}, we fuse the two streams adaptively with the learned gate for both static and dynamic spatial reasoning:
\begin{equation}
\boldsymbol{F}
=
\boldsymbol{\alpha}\odot \boldsymbol{V}
+
(1-\boldsymbol{\alpha})\odot \boldsymbol{G},
\label{eq:ggf_fuse2}
\end{equation}
where $\odot$ denotes an element-wise multiplication. $\boldsymbol{F}$ is the fused feature map, which inherits both visual and geometry cues reasonably under the control of the gate $\boldsymbol{\alpha}$.
Typically, in static spatial reasoning, $\boldsymbol{F}$ will be fed into the VLM backbone directly, together with the prompt tokens, to achieve spatial reasoning. In dynamic spatial reasoning, the overall dynamic geometry information in $\boldsymbol{Z}_{\text{G}}$ (as in Eq.~\eqref{eq:pcm_crossattn_geo}) is also helpful to perceive the global context, which is appended further to $\boldsymbol{F}$ as $[\boldsymbol{F},\boldsymbol{Z}_{\text{G}}]$ and then fed into VLM in practice.
% \begin{equation}
% \boldsymbol{F}=[\boldsymbol{F},\boldsymbol{Z}_{\text{G}}],
% \label{eq:ggf_final2}
% \end{equation}
% and feed $\boldsymbol{F}$ into the VLM backbone in practice.

From Eqs.~\eqref{eq:ggf_crossattn_retrieve} to~\eqref{eq:ggf_fuse2}, Geometry-Guided Fusion adaptively mixes geometric cues with vision tokens using a learned gate $\boldsymbol{\alpha}$.
This design allows geometry features to dominate at locations where they provide useful evidence.

\subsection{Implementation Details}
\label{subsec:details}

This section provides implementation details of \textbf{GeoSR}, including training and inference settings for both static and dynamic spatial reasoning.

\noindent \textbf{Static spatial reasoning.}
% \paragraph{Static Spatial Reasoning.}
For static spatial reasoning, we use Qwen2.5-VL-7B as the backbone VLM~\cite{bai2025qwen25vl} and VGGT~\cite{wang2025vggt} as the pretrained 3D model to extract geometry tokens.
Following our static masking design from Eqs.~\eqref{eq:gum_apply} to~\eqref{eq:gum_random}, we adopt MAE-style random masking with masking ratio $\gamma=0.8$ in Eq.~\eqref{eq:gum_random} and masking enable probability $\beta=0.5$ in Eq.~\eqref{eq:gum_enable}.
During fine-tuning, we update the VLM backbone together with the Geometry-Guided Fusion module, while keeping the tokenizers and the 3D geometry model frozen.
We use the same data splits of SPAR-7M~\cite{zhang2025from} and LLaVA-Hound~\cite{zhang2025llava} as in~\cite{zheng2025learning}.
We train our model for 1 epoch with Adam~\cite{kingma2014adam} optimizer, a batch size of 64, and a learning rate of $1\times 10^{-5}$ with linear warmup for 150 steps followed by cosine decay to 0.
Inference follows training except that masking is disabled and the full vision tokens are fed into the fusion module.

\noindent \textbf{Dynamic spatial reasoning.}
% \paragraph{Dynamic Spatial Reasoning.}
For dynamic spatial reasoning, we keep the Qwen2.5-VL-7B backbone~\cite{bai2025qwen25vl} but switch the 3D model to $\pi^3$~\cite{wang2026pi} to better capture dynamic geometry cues.
We follow the dynamic masking design in Eqs.~\eqref{eq:pcm_bottleneck} to~\eqref{eq:pcm_topk} with bottleneck length $L_{\text{B}}=32$ in Eq.~\eqref{eq:pcm_bottleneck}, masking ratio $\gamma=0.8$ in Eq.~\eqref{eq:pcm_topk}, and enable masking with probability $\beta=0.5$ in Eq.~\eqref{eq:gum_enable}.
The training set is DSR-Train~\cite{zhou2026learning} following its official protocol.
We train the model for 1 epoch with Adam~\cite{kingma2014adam} optimizer, a batch size of 32, and a learning rate of $2\times 10^{-7}$ with linear warmup for 50 steps.
Inference is the same as training except that masking is disabled.

All experiments are conducted on 4$\times$H200 GPUs with 141GB memory each. Training takes about 14 hours for static spatial reasoning with DeepSpeed~\cite{rasley2020deepspeed} ZeRO-2 and about 20 hours for dynamic spatial reasoning with ZeRO-3 Offload.

\section{Experiments}
\label{sec:experiments}

We perform extensive experiments on both static and dynamic spatial reasoning benchmarks to evaluate the effectiveness of \textbf{GeoSR}, proving that it can significantly improve spatial reasoning performance by making geometry matter.
We further conduct ablation studies to validate the contribution of each component and analyze the hyperparameters and the computational cost.

\begin{table}[t]
  \centering
  \small
  \caption{
    Performance on static spatial reasoning.
    \colorbox{blue!27}{Best}, \colorbox{blue!16}{second-best}, and \colorbox{blue!7}{third-best} results are highlighted.
  }
  \setlength{\tabcolsep}{2.0pt}
  \renewcommand{\arraystretch}{1.0}

  % group row
  \newcommand{\grouprow}[1]{%
    \rowcolor{black!10}%
    \multicolumn{10}{c}{\textit{#1}}\\
  }

  % Top-1/2/3 highlighting (blue palette)
  \newcommand{\First}{\cellcolor{blue!27}}
  \newcommand{\Second}{\cellcolor{blue!16}}
  \newcommand{\Third}{\cellcolor{blue!7}}

  \resizebox{\textwidth}{!}{%
  \begin{tabular}{lcccccccccc}
    \toprule
    \multirow[c]{3}{*}{\strut Models} &
    \multicolumn{8}{c}{Subtask Types} &
    \multirow[c]{3}{*}{\strut Avg.} \\
    \cmidrule(lr){2-9}
    & \multicolumn{4}{c}{Numerical Questions}
    & \multicolumn{4}{c}{Multiple-Choice Questions}
    & \\
    \cmidrule(lr){2-5}\cmidrule(lr){6-9}
    & Obj. Cnt.
    & Abs. Dist.
    & Obj. Size
    & Room Size
    & Rel. Dist.
    & Rel. Dir.
    & Route Plan
    & Appr. Order
    & \\
    \midrule

    \grouprow{Proprietary Models (API)}
    \multicolumn{1}{l|}{GPT-4o~\cite{hurst2024gpt}}            & 43.1 & 34.1 & \First 68.6 & \multicolumn{1}{c|}{\First 64.2} & \Third 48.3 & 43.1 & 29.4 & \multicolumn{1}{c|}{51.3} & 47.8 \\
    \multicolumn{1}{l|}{Gemini-1.5 Flash~\cite{team2024gemini}}  & 50.8 & 33.6 & 56.5 & \multicolumn{1}{c|}{45.2} & 48.0 & 39.8 & 32.7 & \multicolumn{1}{c|}{\Third 59.2} & 45.7 \\
    \multicolumn{1}{l|}{Gemini-1.5-Pro~\cite{team2024gemini}}    & 49.6 & 28.8 & \Third 58.6 & \multicolumn{1}{c|}{49.4} & 46.0 & \First 48.1 & \First 42.0 & \multicolumn{1}{c|}{\First 68.0} & \Third 48.8 \\
    % \multicolumn{1}{l|}{Gemini-2.0 Flash~\cite{deepmind_gemini20flash_modelcard_2025}}  & 52.4 & 30.6 & \Second 66.7 & \multicolumn{1}{c|}{31.8} & \First 56.0 & \Second 46.3 & 24.5 & \multicolumn{1}{c|}{55.1} & 45.4 \\
    \midrule

    \grouprow{General Video Understanding Models}
    \multicolumn{1}{l|}{Qwen2.5-VL-7B~\cite{bai2025qwen25vl}}     & 40.9 & 14.8 & 43.4 & \multicolumn{1}{c|}{10.7} & 38.6 & 38.5 & 33.0 & \multicolumn{1}{c|}{29.8} & 33.0 \\
    \multicolumn{1}{l|}{Qwen2.5-VL-72B~\cite{bai2025qwen25vl}}    & 25.1 & 29.3 & 54.5 & \multicolumn{1}{c|}{38.8} & 38.2 & 37.0 & 34.0 & \multicolumn{1}{c|}{28.9} & 37.0 \\
    \multicolumn{1}{l|}{LLaVA-Video-7B~\cite{zhang2025llava}}    & 48.5 & 14.0 & 47.8 & \multicolumn{1}{c|}{24.2} & 43.5 & 42.4 & 34.0 & \multicolumn{1}{c|}{30.6} & 35.6 \\
    \multicolumn{1}{l|}{LLaVA-Video-72B~\cite{zhang2025llava}}   & 48.9 & 22.8 & 57.4 & \multicolumn{1}{c|}{35.3} & 42.4 & 36.7 & 35.0 & \multicolumn{1}{c|}{48.6} & 40.9 \\
    \multicolumn{1}{l|}{InternVL2-8B~\cite{team2024internvl2}}      & 25.7 & 24.0 & 20.0 & \multicolumn{1}{c|}{29.2} & 32.1 & 44.1 & 30.4 & \multicolumn{1}{c|}{6.3}  & 26.5 \\
    \multicolumn{1}{l|}{InternVL3-78B~\cite{zhu2025internvl3}}     & \First 71.2 & \First 53.7 & 44.4 & \multicolumn{1}{c|}{39.5} & \First 55.9 & 39.5 & 28.9 & \multicolumn{1}{c|}{54.5} & 48.4 \\
    \midrule

    \grouprow{Spatial Reasoning Models}
    \multicolumn{1}{l|}{SAT-LLaVA-Video~\cite{ray2025sat}}   & --   & --   & --   & \multicolumn{1}{c|}{47.3} & 41.1 & 37.1 & \Second 36.1 & \multicolumn{1}{c|}{40.4} & --   \\
    \multicolumn{1}{l|}{SPAR~\cite{zhang2025from}}              & --   & --   & --   & \multicolumn{1}{c|}{--}   & --   & --   & --   & \multicolumn{1}{c|}{--}   & 41.1 \\
    \multicolumn{1}{l|}{Spatial-MLLM~\cite{wu2025spatial}}      & 65.3 & 34.8 & \Second 63.1 & \multicolumn{1}{c|}{45.1} & 41.3 & \Second 46.2 & 33.5 & \multicolumn{1}{c|}{46.3} & 48.4 \\
    \multicolumn{1}{l|}{VG-LLM~\cite{zheng2025learning}}            & \Third 67.9 & \Third 37.7 & \Third 58.6 & \multicolumn{1}{c|}{\Third 62.0} & 46.6 & 40.7 & 32.4 & \multicolumn{1}{c|}{\Third 59.2} & \Second 50.7 \\
    \multicolumn{1}{l|}{\textbf{GeoSR (Ours)}}      & \Second 68.3 & \Second 38.7 & 57.4 & \multicolumn{1}{c|}{\Second 62.3} & \Second 48.7 & \Third 44.4 & \Third 35.6 & \multicolumn{1}{c|}{\Second 59.5} & \First 51.9 \\
    \bottomrule
  \end{tabular}%
  }

  \label{tab:static}
\end{table}

\subsection{Static Spatial Reasoning}
\label{subsec:3d}

\noindent \textbf{Benchmarks.}
% \paragraph{Benchmarks.}
We evaluate static spatial reasoning on VSI-Bench~\cite{yang2025thinking}, which contains over 5k QA pairs from 288 real videos.
It focuses on mostly rigid scenes where viewpoint and visibility change across frames, so correct answers often require stable geometric understanding rather than appearance cues.
VSI-Bench includes numerical and multiple-choice questions.
Numerical subtasks cover object counting (\emph{Obj. Cnt.}), absolute distance estimation (\emph{Abs. Dist.}), object size (\emph{Obj. Size}), and room size (\emph{Room Size}).
Multiple-choice subtasks cover relative distance (\emph{Rel. Dist.}), relative direction (\emph{Rel. Dir.}), route planning (\emph{Route Plan}), and appearance ordering (\emph{Appr. Order}).
Following the official protocol, we report accuracy for multiple-choice questions, and mean relative accuracy for numerical questions, which aggregates scores under several relative error tolerances.

\noindent \textbf{Baselines and results.}
% \paragraph{Baselines and Results.}
We compare \textbf{GeoSR} with three groups of baselines: (1) proprietary API models as strong reference~\cite{hurst2024gpt,team2024gemini}, (2) representative open-source video models that are trained for general video understanding~\cite{bai2025qwen25vl,zhang2025llava,team2024internvl2,zhu2025internvl3}, (3) spatial reasoning models that are enhanced by spatial supervision and geometry priors~\cite{ray2025sat,zhang2025from,wu2025spatial,zheng2025learning}.
Overall, as shown in Table~\ref{tab:static}, \textbf{GeoSR} improves over the corresponding baselines under the same evaluation settings, which suggests that geometry tokens become more useful during reasoning rather than remaining auxiliary context.
% We also report subtask-level scores, the improvements are more evident on those that require viewpoint-aware judgment and quantitative estimation, where shortcuts based on texture or category priors are less reliable when viewpoint changes or partial occlusion occurs.

\begin{table}[t]
  \centering
  \small
  \caption{
    Performance on dynamic spatial reasoning.
    \colorbox{blue!27}{Best}, \colorbox{blue!16}{second-best}, and \colorbox{blue!7}{third-best} results are highlighted.  
  }
  \setlength{\tabcolsep}{2.0pt}
  \renewcommand{\arraystretch}{1.0}

  \newcommand{\grouprow}[1]{%
    \rowcolor{black!10}%
    \multicolumn{15}{c}{\textit{#1}}\\
  }

  % Top-1/2/3 highlighting (blue palette)
  \newcommand{\First}{\cellcolor{blue!27}}
  \newcommand{\Second}{\cellcolor{blue!16}}
  \newcommand{\Third}{\cellcolor{blue!7}}

  \resizebox{\textwidth}{!}{%
  \begin{tabular}{lcccccccccccccc}
    \toprule
    \multirow{3}{*}{Models} &
    \multicolumn{13}{c}{Subtask Types} &
    \multirow{3}{*}{Avg.} \\
    \cmidrule(lr){2-14}
    & \multicolumn{6}{c}{Absolute}
    & \multicolumn{6}{c}{Relative}
    & \multirow{2}{*}{\makecell{Non.\\Temp.}} & \\
    \cmidrule(lr){2-7}\cmidrule(lr){8-13}
    & Dist.
    & Dir.
    & Ori.
    & Spd.
    & \makecell{Spd.\\Comp.}
    & \makecell{Dir.\\Pred.}
    & Dist.
    & Dir.
    & Ori.
    & Spd.
    & \makecell{Spd.\\Comp.}
    & \makecell{Dir.\\Pred.}
    & & \\
    \midrule

    \grouprow{Proprietary Models (API)}
    \multicolumn{1}{l|}{GPT-4o~\cite{hurst2024gpt}}           & 18.8 & 29.2 & 26.8 & 29.7 & 21.5 & \multicolumn{1}{c}{26.2} & 24.1 & 23.8 & 17.2 & \Third 32.3 & 22.8 & \multicolumn{1}{c}{24.4} & \multicolumn{1}{c|}{34.7} & 26.4 \\
    \multicolumn{1}{l|}{GPT-5~\cite{singh2025openai}}            & 21.1 & 41.5 & 48.7 & 34.5 & 33.3 & \multicolumn{1}{c}{\Third 34.7} & 17.2 & \Third 44.3 & 41.9 & 21.2 & 25.0 & \multicolumn{1}{c}{30.9} & \multicolumn{1}{c|}{26.7} & 30.8 \\
    \multicolumn{1}{l|}{Gemini-2.5-Flash~\cite{comanici2025gemini}} & 18.8 & 27.6 & 19.5 & 25.0 & 23.6 & \multicolumn{1}{c}{22.0} & 11.2 & 28.4 & 30.8 & 23.2 & 17.1 & \multicolumn{1}{c}{22.6} & \multicolumn{1}{c|}{\Third 38.8} & 24.9 \\
    \multicolumn{1}{l|}{Gemini-2.5-Pro~\cite{comanici2025gemini}}   & 20.0 & \Third 44.6 & 53.6 & 27.3 & 38.7 & \multicolumn{1}{c}{30.5} & 23.2 & 32.9 & \Third 43.2 & 17.1 & 28.5 & \multicolumn{1}{c}{27.9} & \multicolumn{1}{c|}{34.3} & 31.7 \\
    \midrule

    \grouprow{General Video Understanding Models}
    \multicolumn{1}{l|}{Qwen2.5-VL-7B~\cite{bai2025qwen25vl}}             & 18.8 & 15.3 & 14.6 & 42.8 & 29.0 & \multicolumn{1}{c}{19.4} & 31.8 & 19.3 & 11.1 & 22.2 & 19.2 & \multicolumn{1}{c}{20.2} & \multicolumn{1}{c|}{30.1} & 23.5 \\
    \multicolumn{1}{l|}{Qwen2.5-VL-32B~\cite{bai2025qwen25vl}}            & 31.7 & 21.5 & 23.1 & 44.0 & 36.5 & \multicolumn{1}{c}{25.4} & 27.5 & 23.8 & 37.0 & 27.2 & 29.2 & \multicolumn{1}{c}{21.4} & \multicolumn{1}{c|}{36.2} & 29.9 \\
    \multicolumn{1}{l|}{Qwen3-VL-8B-Instruct~\cite{bai2025qwen3}}      & 23.5 & 24.6 & 42.6 & 29.7 & 27.9 & \multicolumn{1}{c}{33.8} & 18.1 & 28.4 & 34.5 & 24.2 & 22.1 & \multicolumn{1}{c}{27.9} & \multicolumn{1}{c|}{33.5} & 28.7 \\
    \multicolumn{1}{l|}{Qwen3-VL-30B-A3B-Instruct~\cite{bai2025qwen3}} & 25.8 & 27.6 & 46.3 & 30.9 & 31.1 & \multicolumn{1}{c}{\Third 34.7} & 20.6 & 29.5 & 37.0 & 28.2 & 24.2 & \multicolumn{1}{c}{\Third 31.5} & \multicolumn{1}{c|}{35.4} & 31.1 \\
    \multicolumn{1}{l|}{LLaVA-Video-7B~\cite{zhang2025llava}}            & 22.3 & 16.9 & 25.6 & 33.3 & 45.1 & \multicolumn{1}{c}{24.5} & 24.1 & 15.9 & 17.2 & 19.1 & 24.2 & \multicolumn{1}{c}{21.4} & \multicolumn{1}{c|}{33.9} & 25.9 \\
    \multicolumn{1}{l|}{VideoRefer~\cite{yuan2025videorefer}}                & 23.5 & 18.4 & 25.6 & 33.5 & 45.4 & \multicolumn{1}{c}{27.1} & 25.0 & 16.1 & 18.5 & 20.2 & 26.4 & \multicolumn{1}{c}{22.6} & \multicolumn{1}{c|}{34.7} & 26.9 \\
    \multicolumn{1}{l|}{InternVL3.5-8B~\cite{wang2025internvl3}}            & 23.5 & 27.6 & 28.0 & 34.5 & 24.7 & \multicolumn{1}{c}{27.9} & 22.4 & 17.0 & 19.7 & 28.2 & 30.0 & \multicolumn{1}{c}{14.2} & \multicolumn{1}{c|}{30.1} & 25.4 \\
    \multicolumn{1}{l|}{InternVL3.5-38B~\cite{wang2025internvl3}}           & 25.8 & 27.8 & 29.2 & 34.2 & 24.7 & \multicolumn{1}{c}{28.5} & 26.7 & 16.3 & 23.4 & 29.2 & \Third 32.1 & \multicolumn{1}{c}{15.4} & \multicolumn{1}{c|}{31.3} & 26.7 \\
    \midrule

    \grouprow{Spatial Reasoning Models}
    \multicolumn{1}{l|}{VLM-3R~\cite{fan2026vlm}}           & 28.2 & 27.6 & 31.7 & 42.8 & 38.7 & \multicolumn{1}{c}{33.0} & 34.4 & 23.8 & 30.8 & 22.2 & 26.4 & \multicolumn{1}{c}{29.1} & \multicolumn{1}{c|}{35.0} & 31.4 \\
    \multicolumn{1}{l|}{VG-LLM~\cite{zheng2025learning}}           & \Third 55.2 & 32.3 & \Third 58.5 & \Third 57.1 & \Third 51.6 & \multicolumn{1}{c}{32.2} & \Third 56.0 & 36.3 & 32.0 & 30.3 & \Third 32.1 & \multicolumn{1}{c}{29.1} & \multicolumn{1}{c|}{27.9} & \Third 38.4 \\
    \multicolumn{1}{l|}{GSM~\cite{zhou2026learning}}              & \Second 87.0 & \Second 73.8 & \Second 84.1 & \Second 73.8 & \Second 72.0 & \multicolumn{1}{c}{\Second 35.5} & \Second 75.8 & \Second 76.1 & \Second 77.7 & \Second 60.6 & \Second 37.1 & \multicolumn{1}{c}{\Second 35.1} & \multicolumn{1}{c|}{\Second 46.4} & \Second 58.9 \\
    \multicolumn{1}{l|}{\textbf{GeoSR (Ours)}}     & \First 88.0 & \First 84.4 & \First 87.8 & \First 83.3 & \First 78.5 & \multicolumn{1}{c}{\First 41.0} & \First 79.5 & \First 83.0 & \First 83.8 & \First 76.5 & \First 48.9 & \multicolumn{1}{c}{\First 39.8} & \multicolumn{1}{c|}{\First 52.9} & \First 66.1 \\
    \bottomrule
  \end{tabular}%
  }
  \label{tab:dynamic}
\end{table}

\subsection{Dynamic Spatial Reasoning}
\label{subsec:4d}

\noindent \textbf{Benchmarks.}
% \paragraph{Benchmarks.}
We evaluate dynamic spatial reasoning on DSR-Bench~\cite{zhou2026learning}, which contains 1484 QA pairs from 575 in-the-wild videos with evolving dynamic spatial relations due to object and camera motion.
Correct answers require spatiotemporal consistency rather than static cues.
DSR-Bench is organized into absolute, relative, and non-template question types.
Absolute types include distance (\emph{Dist.}), direction (\emph{Dir.}), orientation (\emph{Ori.}), and speed (\emph{Spd.}), and also include comparison and prediction variants such as speed comparison (\emph{Spd. Comp.}) and direction prediction (\emph{Dir. Pred.}).
Relative types ask about relations defined with respect to another entity, another time point, or another reference frame, including similar types to the absolute ones under such references.
Beyond these structured categories, \emph{Non-Temp.} contains questions with more flexible phrasing that are not restricted to fixed templates.
We follow the official evaluation protocol in~\cite{zhou2026learning} and report accuracy for each type and the overall average.

\noindent \textbf{Baselines and results.}
% \paragraph{Baselines and Results.}
Similar to the static spatial reasoning settings, Table~\ref{tab:dynamic} compares \textbf{GeoSR} with proprietary API models~\cite{hurst2024gpt,singh2025openai,comanici2025gemini}, general video understanding models~\cite{bai2025qwen25vl,bai2025qwen3,zhang2025llava,yuan2025videorefer,wang2025internvl3} and spatial reasoning models~\cite{fan2026vlm,zheng2025learning,zhou2026learning}.
% This comparison is important because strong video models can capture rich temporal semantics, while spatial reasoning models are designed to strengthen geometry-related capability, and DSR-Bench requires both temporal consistency and spatial correctness.
Results in Table~\ref{tab:dynamic} show that DSR-Bench is challenging for general models while spatial reasoning models perform significantly better, which suggests that geometry cues are particularly helpful for dynamic spatial reasoning.
Furthermore, \textbf{GeoSR} rivals the best-performing model and achieves the highest score on each subtask type, which demonstrates that the proposed design highly benefits dynamic spatial reasoning by unleashing geometry cues and adaptively token fusing.
% Across subtask types, \textbf{GeoSR} improves over the base model, showing that the proposed design remains effective when spatial evidence is dynamic and appearance cues become less stable under motion and viewpoint changes.
% The gains are particularly meaningful on categories that depend on how relations change over time, such as direction prediction and relative comparisons under motion, where relying mainly on weak 2D evidence often breaks when the motion is continuous or when occlusion occurs.

\begin{table}[t]
  \centering
  \small
  \caption{Ablations on \textbf{GeoSR} components for static spatial reasoning.}
  \setlength{\tabcolsep}{4.0pt}
  \renewcommand{\arraystretch}{1.0}

  \resizebox{\textwidth}{!}{%
  \begin{tabular}{ccccccccccccc}
    \toprule
    \multirow{2}{*}{Setting} &
    \multicolumn{3}{c}{Module} &
    \multicolumn{8}{c}{Subtask Types} &
    \multirow{2}{*}{Avg.} \\
    \cmidrule(lr){2-4}\cmidrule(lr){5-12}
    & \makecell{Geo.\\Mask.}
    & \makecell{Geo.\\Fus.}
    & \makecell{Ori.\\Fus.}
    & \makecell{Obj.\\Cnt.}
    & \makecell{Abs.\\Dist.}
    & \makecell{Obj.\\Size}
    & \makecell{Room\\Size}
    & \makecell{Rel.\\Dist.}
    & \makecell{Rel.\\Dir.}
    & \makecell{Route\\Plan}
    & \makecell{Appr.\\Order}
    & \\
    \midrule

    (a) & $\checkmark$ & $\checkmark$ &            & 68.3 & 38.7 & 57.4 & 62.3 & 48.7 & 44.4 & 35.6 & 59.5 & 51.9 \\
    (b) & $\checkmark$ &              & $\checkmark$ & 67.7 & 38.5 & 57.7 & 61.6 & 46.5 & 40.9 & 34.0 & 52.9 & 50.0 \\
    (c) & $\checkmark$ &              &            & 68.1 & 38.8 & 57.7 & 60.5 & 46.8 & 42.1 & 26.9 & 55.7 & 49.6 \\
    (d) &            & $\checkmark$ &            & 68.2 & 38.0 & 58.7 & 57.6 & 45.6 & 42.5 & 34.5 & 61.8 & 50.9 \\
    (e) &            &              & $\checkmark$ & 67.2 & 37.8 & 58.0 & 60.9 & 44.9 & 41.8 & 34.0 & 56.6 & 50.2 \\
    (f) &            &              &            & 68.6 & 36.0 & 57.9 & 59.7 & 45.8 & 38.9 & 30.4 & 60.2 & 49.8 \\
    % (f) &            &              &            & 69.7 & 36.3 & 57.5 & 57.5 & 45.1 & 34.4 & 30.4 & 61.5 & 49.0 \\
    \bottomrule
  \end{tabular}%
  }

  \label{tab:ablation_modules}
\end{table}

\begin{table}[t]
  \centering
  % \small
  \caption{Ablations on \textbf{GeoSR} components for dynamic spatial reasoning.}
  \setlength{\tabcolsep}{1.0pt}
  \renewcommand{\arraystretch}{1.0}

  \resizebox{\textwidth}{!}{%
  \begin{tabular}{cccccccccccccccccc}
    \toprule
    \multirow{2}{*}{Setting} &
    \multicolumn{3}{c}{Module} &
    \multicolumn{13}{c}{Subtask Types} &
    \multirow{2}{*}{Avg.} \\
    \cmidrule(lr){2-4}\cmidrule(lr){5-17}
    & \makecell{Geo.\\Mask.}
    & \makecell{Geo.\\Fus.}
    & \makecell{Ori.\\Fus.}
    & \makecell{Abs.\\Dist.}
    & \makecell{Abs.\\Dir.}
    & \makecell{Abs.\\Ori.}
    & \makecell{Abs.\\Spd.}
    & \makecell{Abs.\\Spd.\\Comp.}
    & \makecell{Abs.\\Dir.\\Pred.}
    & \makecell{Rel.\\Dist.}
    & \makecell{Rel.\\Dir.}
    & \makecell{Rel.\\Ori.}
    & \makecell{Rel.\\Spd.}
    & \makecell{Rel.\\Spd.\\Comp.}
    & \makecell{Rel.\\Dir.\\Pred.}
    & \makecell{Non.\\Temp.}
    & \\
    \midrule

    (a) & $\checkmark$ & $\checkmark$ &            & 88.0 & 84.4 & 87.8 & 83.3 & 78.5 & 41.0 & 79.5 & 83.0 & 83.8 & 76.5 & 48.9 & 39.8 & 52.9 & 66.1 \\
    (b) & $\checkmark$ &              & $\checkmark$ & 86.7 & 81.3 & 87.8 & 78.6 & 78.5 & 41.9 & 78.6 & 84.1 & 83.8 & 75.5 & 43.9 & 34.9 & 53.8 & 64.7 \\
    (c) & $\checkmark$ &              &            & 86.7 & 67.2 & 80.5 & 76.2 & 67.7 & 40.2 & 77.7 & 75.0 & 71.3 & 55.1 & 36.0 & 32.5 & 48.7 & 58.1 \\
    (d) &            & $\checkmark$ &            & 88.0 & 84.4 & 89.0 & 77.4 & 78.5 & 41.9 & 77.7 & 87.5 & 82.5 & 70.4 & 41.7 & 38.6 & 54.2 & 64.9 \\
    (e) &            &              & $\checkmark$ & 88.0 & 78.1 & 86.6 & 75.0 & 76.3 & 40.2 & 75.9 & 79.5 & 81.3 & 69.4 & 41.7 & 35.5 & 53.4 & 62.8 \\
    (f) &            &              &            & 86.7 & 79.7 & 85.3 & 75.0 & 79.6 & 41.0 & 75.0 & 83.0 & 82.5 & 74.5 & 43.9 & 39.8 & 51.7 & 64.0 \\
    \bottomrule
  \end{tabular}%
  }

  \label{tab:ablation_modules_dynamic}
\end{table}

\subsection{Ablation Studies}
\label{subsec:ablation}

We conduct ablations on both static (Table~\ref{tab:ablation_modules}) and dynamic (Table~\ref{tab:ablation_modules_dynamic}) benchmarks to assess the effects of \textbf{Geometry-Unleashing Masking} (Geo. Mask.), \textbf{Geometry-Guided Fusion} (Geo. Fus.), and the original naive fusion used in prior geometry-aware frameworks (Ori. Fus.).
We take the full \textbf{GeoSR} model (a) as the reference.
Replacing Geo. Fus. with the Ori. Fus. (b) reduces the performance on both benchmarks, indicating that naive fusion is less effective at turning geometry tokens into actionable evidence.
Removing Geo. Fus. and Ori. Fus. together while keeping Geo. Mask. (c) further degrades the results, suggesting that masking alone cannot fully exploit geometry without a well-designed mechanism to route geometry cues into the VLM backbone.
Conversely, removing Geo. Mask. but keeping Geo. Fus. (d) also leads to a noticeable drop compared to (a), showing that the model can still underutilize geometry without shortcut suppression.
Retaining only Ori. Fus. (e) yields an inferior variant to (d), again highlighting the benefit of adaptive routing over the original fusion.
Finally, we examine the no-geometry model (f).
On the static benchmark, the performance drop from (e) to (f) is relatively moderate, suggesting that for static scenes, the gains from directly injecting geometry can be limited.
On the dynamic benchmark, (f) even outperforms (e), indicating that simple geometry token injection may yield negative effects when geometry cues are fused without proper control, consistent with our motivation in the introduction.
Overall, Geo. Mask. makes geometry \emph{effective} by weakening appearance shortcuts during training, while Geo. Fus. makes geometry usage \emph{reasonable} via adaptive routing.
The improvements are more pronounced on the dynamic benchmark, where motion and occlusion make appearance cues less stable and thus benefit more from enforced geometry reliance and fine-grained fusion.

\subsection{Analysis}
\label{subsec:analysis}

\begin{table}[t]
  \centering
  \small

  \begin{minipage}[t]{0.43\linewidth}
    \centering
    \small
    \setlength{\tabcolsep}{3pt}
    \renewcommand{\arraystretch}{1.0}
    \caption{Hyperparameter analysis. The best configuration is \textbf{bolded}.}
    \label{tab:hyperparam}
    \begin{tabular}{lccc}
      \toprule
      $\gamma$ & \textbf{0.8} & 0.6 & 0.4 \\
      Avg.     & \textbf{66.1} & 65.9 & 65.0 \\
      \midrule
      $\beta$  & 0.7 & \textbf{0.5} & 0.3 \\
      Avg.     & 64.5 & \textbf{66.1} & 64.9 \\
      \bottomrule
    \end{tabular}
  \end{minipage}
  % \hfill
  \begin{minipage}[t]{0.55\linewidth}
    \centering
    \small
    \setlength{\tabcolsep}{2pt}
    \renewcommand{\arraystretch}{1.0}
    \caption{Computational overhead. Model size includes the VLM backbone and all embedders.}
    \label{tab:compute}
    \begin{tabular}{lccc}
      \toprule
      Models & Time & Size & Mem. \\
      \midrule
      Qwen2.5-VL-7B~\cite{bai2025qwen25vl}       & 0.37s & 8.76B & 18.04GB \\
      w/ Geo. (Baseline)  & 0.40s & 9.16B & 18.81GB \\
      \textbf{GeoSR (Ours)}        & 0.41s & 9.23B & 18.95GB \\
      \bottomrule
    \end{tabular}
  \end{minipage}
\end{table}

\noindent \textbf{Hyperparameter settings.}
We validate the masking hyperparameter settings of $\gamma$ in Eq.~\eqref{eq:pcm_topk} and $\beta$ in Eq.~\eqref{eq:gum_enable} on DSR-Bench~\cite{zhou2026learning}.
Table~\ref{tab:hyperparam} shows that small $\gamma$ and $\beta$ leave appearance shortcuts largely intact and bring limited gains, while overly large $\beta$ removes too much context and can hurt stability.
We find that $\gamma=0.8$ and $\beta=0.5$ strike a good balance.
% Overall, \textbf{GeoSR} is robust to moderate choices of $(\gamma,\beta)$ and performs consistently well under a balanced setting.

\noindent \textbf{Computational overhead.}
We report runtime, model size, and peak memory for dynamic reasoning on DSR-Bench~\cite{zhou2026learning} in Table~\ref{tab:compute}, testing on a single H200 GPU.
% \textbf{GeoSR} adds a lightweight fusion pathway on top of a frozen geometry encoder and a VLM backbone.
The extra parameters of \textbf{GeoSR} mainly come from the geometry projection and the gating module, so the parameter increase is minor relative to the geometry-aware baseline.
\textbf{GeoSR} also introduces a negligible runtime increase, showing that the proposed design is efficient and does not require heavy computation to unleash geometry cues and boost the spatial reasoning performance.

\section{Conclusion}
\label{sec:conclusion}
In this paper, we study a failure mode in existing geometry-aware VLMs for spatial reasoning: under the common recipe of naive token fusion and standard fine-tuning, injected geometry tokens can be largely underutilized, bringing limited gains and even harming performance. This suggests that geometry priors, while informative, are not automatically treated as actionable evidence by the VLM backbone.
To address this issue, we propose \textbf{GeoSR}, a new framework that makes geometry matter by encouraging \emph{effective} and \emph{reasonable} geometry usage. \textbf{Geometry-Unleashing Masking} weakens vision-driven shortcuts during training and forces the model to consult geometry tokens, while \textbf{Geometry-Guided Fusion} adaptively routes geometry evidence via a fine-grained gate so that geometry contributes more when it is informative and necessary. Extensive experiments on both static and dynamic spatial reasoning benchmarks validate the effectiveness of these designs, with consistent improvements over prior baselines, especially in dynamic settings.

% \section*{Acknowledgements}
% Please insert your acknowledgments here.

\bibliographystyle{splncs04}
\bibliography{main}

\appendix
\section*{\appendixname}

\section{Qualitative Results}
\label{sec:qualitative}
\begin{figure}[t]
    \centering
    \includegraphics[width=0.99\linewidth]{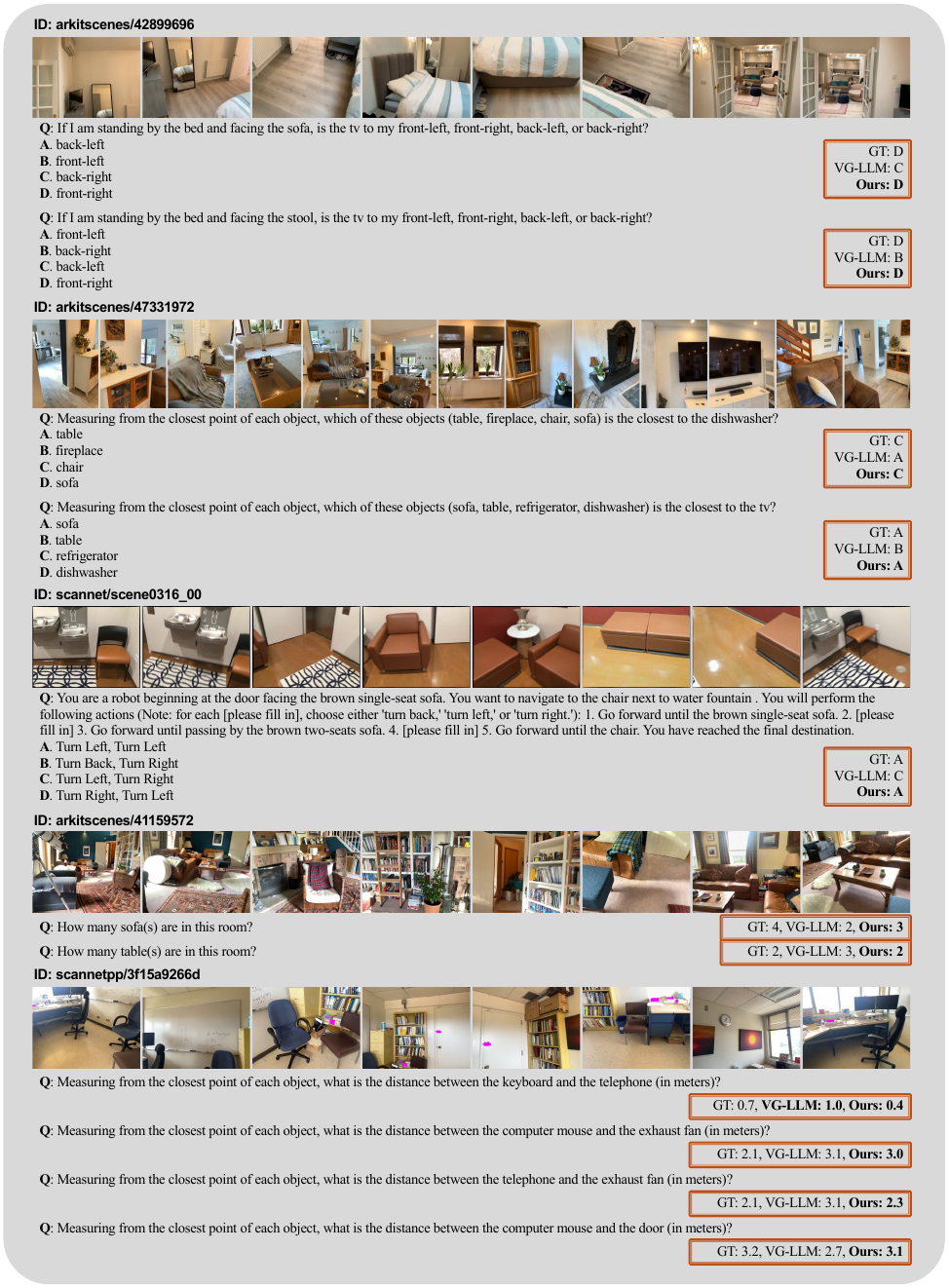}
    \caption{
      Visualization of the static spatial reasoning results on VSI-Bench~\cite{yang2025thinking}.
    }
    \label{fig:static}
\end{figure}

\begin{figure}[t]
    \centering
    \includegraphics[width=0.99\linewidth]{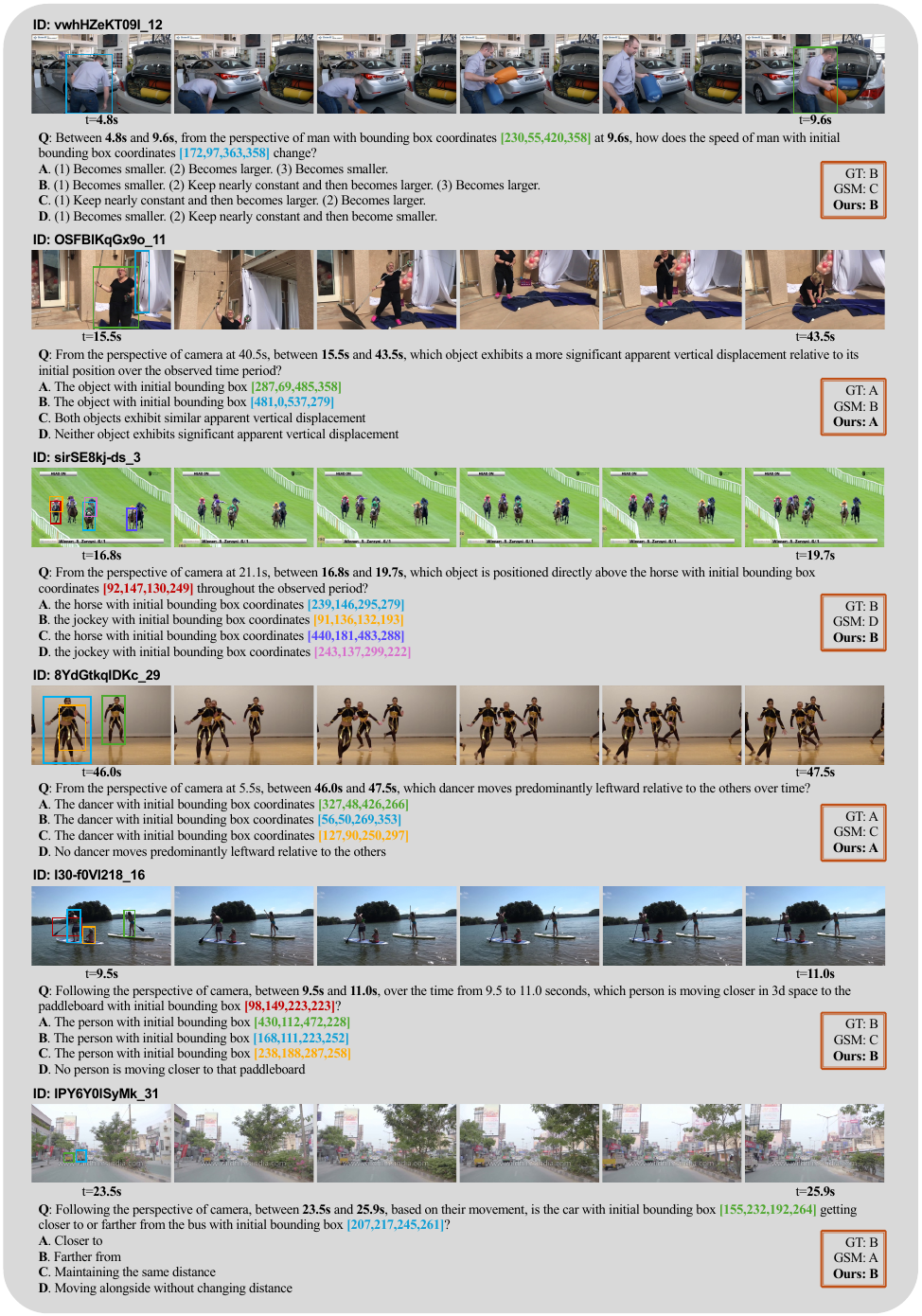}
    \caption{
      Visualization of the dynamic spatial reasoning results on DSR-Bench~\cite{zhou2026learning}.
    }
    \label{fig:dynamic}
\end{figure}

We present qualitative results of \textbf{GeoSR} on both static and dynamic spatial reasoning benchmarks~\cite{yang2025thinking,zhou2026learning}. The examples are shown in Figures~\ref{fig:static} and~\ref{fig:dynamic}, together with comparisons to strong baselines, including VG-LLM~\cite{zheng2025learning} and GSM~\cite{zhou2026learning}. Each example is annotated with its video ID, and the corresponding videos are provided in the supplementary video file.

\begin{figure}[!t]
    \centering
    \includegraphics[width=0.99\linewidth]{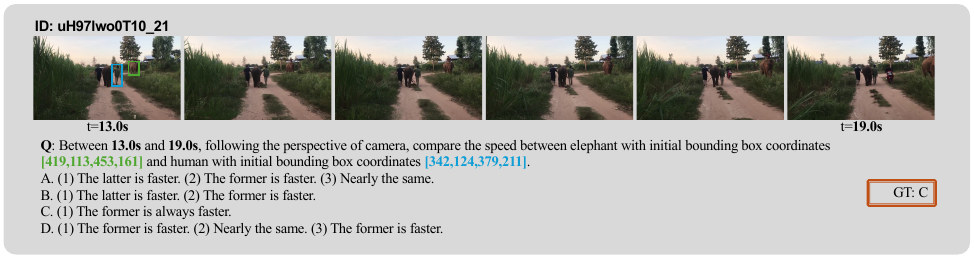}
    \caption{
      Example illustrating ambiguity in the benchmark. The question asks for a speed comparison between the elephant and the person, but the visual evidence is ambiguous from a geometric perspective. The ground-truth answer is ``(1) The latter is faster. (2) The former is faster.'' However, in the video segment from 13.0s to 19.0s, it is difficult to determine which one is faster, since the person and the elephant appear to move at similar speeds.
    }
    \label{fig:case1}
\end{figure}

\begin{figure}[!t]
    \centering
    \includegraphics[width=0.99\linewidth]{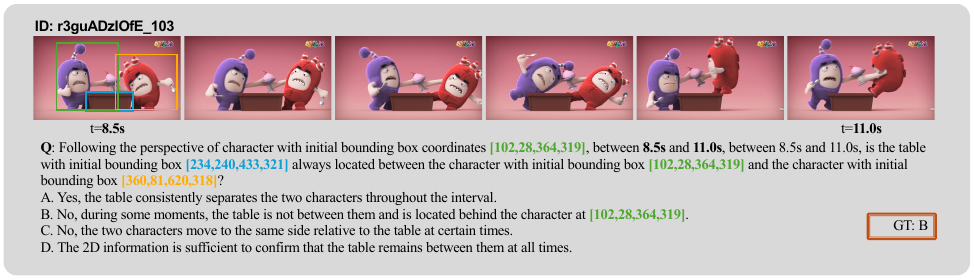}
    \caption{
      Example illustrating ambiguity in the benchmark. The question asks about the relative spatial relation among the table and the two cartoon characters, and the ground-truth answer states that ``No, during some moments, the table is not between them and is located behind the character at [102,28,364,319].'' However, in the video segment from 8.5s to 11.0s, this relation remains difficult to judge. Although the purple character moves toward the table, the character is partially occluded by it, making it unclear whether the table is indeed behind the character.
    }
    \label{fig:case2}
\end{figure}

\section{More Analysis}
\label{sec:analysis}
In the main paper, following prior meticulous architecture in~\cite{wu2025spatial,zheng2025learning,zhou2026learning}, the baseline for dynamic spatial reasoning adopts a QFormer-style architecture, whereas the static one does not. This design choice is reasonable because dynamic spatial reasoning requires aggregating motion patterns and temporal relations across frames, for which QFormer-style query tokens provide an effective mechanism to summarize global video dynamics.
To examine whether such a design is also beneficial for static spatial reasoning, we additionally implement a QFormer-style variant of our \textbf{GeoSR}. As shown in Table~\ref{tab:qformer_ablation}, introducing QFormer does not improve performance. This suggests that, unlike dynamic scenarios, static spatial reasoning relies less on explicit global temporal aggregation, and its main challenge lies more in grounding spatial judgments on geometric cues than in modeling long-range dynamics.

\begin{table}[h!]
  \centering
  \small
  \caption{Performance of GeoSR and GeoSR with QFormer on static spatial reasoning.}
  \setlength{\tabcolsep}{2.0pt}
  \renewcommand{\arraystretch}{1.0}

  \resizebox{\textwidth}{!}{%
  \begin{tabular}{lccccccccc}
    \toprule
    \multirow[c]{3}{*}{\strut Models} &
    \multicolumn{8}{c}{Subtask Types} &
    \multirow[c]{3}{*}{\strut Avg.} \\
    \cmidrule(lr){2-9}
    & \multicolumn{4}{c}{Numerical Questions}
    & \multicolumn{4}{c}{Multiple-Choice Questions}
    & \\
    \cmidrule(lr){2-5}\cmidrule(lr){6-9}
    & Obj. Cnt.
    & Abs. Dist.
    & Obj. Size
    & Room Size
    & Rel. Dist.
    & Rel. Dir.
    & Route Plan
    & Appr. Order
    & \\
    \midrule
    GeoSR     & 68.3 & 38.7 & 57.4 & 62.3 & 48.7 & 44.4 & 35.6 & 59.5 & 51.9 \\
    +QFormer  & 68.6 & 38.8 & 57.8 & 60.6 & 49.0 & 43.5 & 35.6 & 60.2 & 51.8 \\
    \bottomrule
  \end{tabular}%
  }
  \label{tab:qformer_ablation}
\vspace{-4mm}
\end{table}

\section{Limitations and Future Work}
\label{sec:limitations}
\textbf{GeoSR} focuses on model-side improvements that better exploit geometric cues for spatial reasoning on existing benchmarks~\cite{yang2025thinking,zhou2026learning}. Nevertheless, the quality of current datasets may also limit further progress. Since both training and evaluation data are constructed through automatic or semi-automatic pipelines, some question formulations can be ambiguous from a geometric perspective, and some annotations may not perfectly align with the underlying visual evidence. Figures~\ref{fig:case1} and~\ref{fig:case2} provide two representative examples. Such issues may affect the performance of all methods. An important future direction is therefore to improve dataset quality, especially in terms of geometry-aware question construction and annotation consistency, alongside continued advances in model design.

\end{document}